\documentclass[conference]{IEEEtran}
\IEEEoverridecommandlockouts

\usepackage{amsmath,graphicx}

\usepackage{amsmath}
\usepackage{amssymb}
\usepackage{comment}
\usepackage{hyperref}
\usepackage{float}
\usepackage{subcaption}

\title{Out-of-Distribution Radar Detection in Compound Clutter and Thermal Noise through Variational Autoencoders}

\author{
\IEEEauthorblockN{
Y. A. Rouzoumka\IEEEauthorrefmark{1}\IEEEauthorrefmark{2}, 
E. Terreaux\IEEEauthorrefmark{1},
C. Morisseau\IEEEauthorrefmark{1},
J.-P. Ovarlez\IEEEauthorrefmark{1}\IEEEauthorrefmark{2} and
C. Ren\IEEEauthorrefmark{2}
}
\IEEEauthorblockA{
\IEEEauthorrefmark{1}DEMR, ONERA, Universit\'e Paris Saclay, F-91123 Palaiseau, France, \\
\IEEEauthorrefmark{2}SONDRA, CentraleSupélec,  Université Paris-Saclay, 91190 Gif-sur-Yvette, France.
}
}

\begin{document}
%\ninept
%
\maketitle
\begin{abstract}
This paper presents a novel approach to radar target detection using Variational AutoEncoders (VAEs). Known for their ability to learn complex distributions and identify out-of-distribution samples, the proposed VAE architecture effectively distinguishes radar targets from various noise types, including correlated Gaussian and compound Gaussian clutter, often combined with additive white Gaussian thermal noise. Simulation results demonstrate that the proposed VAE outperforms classical adaptive detectors such as the Matched Filter and the Normalized Matched Filter, especially in challenging noise conditions, highlighting its robustness and adaptability in radar applications.
\end{abstract}


\section{Introduction}
\label{sec:intro}

Radar target detection is a fundamental challenge in signal processing \cite{Greco16}, where the goal is to accurately identify the presence of a target within a noisy environment. Traditional detection methods, such as the Matched Filter (MF), Normalized Matched Filter (NMF), and their adaptive variants (AMF \cite{Robey1992ACA}, Kelly \cite{4104190} and ANMF \cite{301849}), have been extensively studied and are well-known for their effectiveness in scenarii with Gaussian noise. However, in real-world applications, radar signals often encounter more complex noise structures, including compound Gaussian noise plus additive white Gaussian thermal noise, which significantly degrade the performance of these classical detectors in terms of probability detection and false alarm regulation.

Recent advancements in machine learning, particularly in the domain of Deep Learning \cite{GoodBengCour16}, offer promising alternatives for enhancing radar detection capabilities. Among these, VAEs have emerged as powerful tools for Out-Of-Distribution (OOD) detection \cite{ran2021, yang2024} by modeling the underlying probability distribution of input data \cite{Kingma_2019}.  Traditional approaches to anomaly detection  \cite{MRAMJ2022} often struggle with complex environmental noise, but VAE-based methods offer robustness by leveraging the ability to detect events in data that lies outside the distribution learned during training. This is particularly relevant for radar applications, where environmental variability and noise complexity are common. While VAE-based anomaly detection has demonstrated success in various fields, including acoustic signal processing \cite{BS2023}, medical imaging \cite{marimont2020},  discharge in high-voltage machines \cite{MSWPNM2021}, its application to radar target detection remains relatively unexplored. For instance, VAE architectures have shown promise in detecting human body motion using frequency-modulated continuous waves radar \cite{KSS2023}.

In this paper, we explore the potential of VAEs for radar target detection, focusing on their ability to handle various clutter and noise models that pose significant challenges to traditional methods. Our approach integrates OOD detection techniques to ensure that targets are distinguishable in complex heterogeneous environments. The paper is structured as follows: Section \ref{sec:statmodel} reviews the statistical models and classical detectors commonly used in radar target detection. Section \ref{sec:deeplearningdetect} introduces the proposed VAE-based detection approach, detailing its architecture, and detection strategy. Section \ref{sec:results} presents the VAE training process and the simulation results comparing the VAE's performance against traditional detectors under various noise conditions. Finally, Section \ref{sec:conclu} concludes the paper, highlighting the advantages of VAEs in radar target detection and potential areas for future research.\\

\indent \textit{Notations}: Matrices are in bold and capital, vectors in bold. For any matrix $\mathbf{A}$ or vector, $\mathbf{A}^T$ is the transpose of $\mathbf{A}$ and $\mathbf{A}^H$ is the Hermitian transpose of $\mathbf{A}$. $\mathbf{I} $ is the identity matrix. $\mathcal{N}(\boldsymbol{\mu},\boldsymbol{\Gamma})$ and $\mathcal{CN}(\boldsymbol{\mu},\boldsymbol{\Gamma})$ are respectively real and complex circular Normal distribution of mean $\boldsymbol{\mu}$ and covariance matrix $\boldsymbol{\Gamma}$. The matrix operator $ \boldsymbol{\mathcal{T}}(.)$ is the Toeplitz matrix operator $\rho \rightarrow \left\{\boldsymbol{\mathcal{T}}(\rho)\right\}_{i,j} = \rho^{|i-j|}$. The symbol $\odot$ denotes the Hadamard element-wise product.

\section{Statistical Model}
\label{sec:statmodel}

\subsection{Hypothesis Testing and Signal Model}
\label{ssec:hypothesis}
In adaptive radar detection, the main problem consists of detecting a complex signal $\alpha \, \mathbf{p} \in \mathbb{C}^m$ corrupted by an additive clutter noise $\mathbf{c}$ and thermal white Gaussian noise vector $\mathbf{n}$ with covariance matrix $\sigma^2 \, \mathbf{I}$, independent of the clutter $\mathbf{c}$. In the case of point-like target, this problem can be stated as the following binary hypothesis test: $\left\{
\begin{array}{ll}
H_0 : \mathbf{z} = \mathbf{c} + \mathbf{n},\,  \\
H_1 : \mathbf{z} = \alpha\,\mathbf{p} + \mathbf{c} + \mathbf{n}, 
\end{array}
\right.$
where $\mathbf{z}$ is the complex $m$-vector of the received signal, $\alpha$ is an unknown complex target amplitude, $\mathbf{p}$ stands for a known {\it steering vector}. 
In homogeneous clutter, $\mathbf{c}$ is modeled by a complex circular Gaussian vector distributed as $\mathbb{C}\mathcal{N}\left(\mathbf{0},\boldsymbol{\Sigma}_c \right)$. In heterogeneous clutter, compound Gaussian model is used instead, $\mathbf{c} = \sqrt{\tau} \, \mathbf{g}$, distributed as $\mathbb{C}\mathcal{N}\left(\mathbf{0}, \tau\,\boldsymbol{\Sigma}_c \right)$ conditionally to the texture $\tau$ $\in \mathbb{R}^+$. The latter represents the power fluctuation from one radar cell to another one. For the sake of simplicity, the average power fluctuation is assumed to be $\mathbb{E}\left[\tau\right] = 1$.

The Signal-to-Noise Ratio (SNR) under hypothesis \(H_1\), after whitening, is defined as $\text{SNR} = |\alpha|^2 \, \mathbf{p}^H \, \boldsymbol{\Sigma}^{-1} \, \mathbf{p}$, 
where $\boldsymbol{\Sigma} = \boldsymbol{\Sigma}_c + \sigma^2 \,\mathbf{I}$. In the following, the power ratio $r= \mathrm{Tr}(\boldsymbol{\Sigma}_c)/(m \, \sigma^2)$ will be fixed to one between clutter and thermal noise powers. 

Without the thermal noise term, the two previous cases have their corresponding optimal solutions (benchmarks) as well as their two adaptive versions. In homogeneous Gaussian environment, the MF leads to:
\begin{equation}
\Lambda_{MF}(\mathbf{z}) = \frac{\left|\mathbf{p}^H \, \boldsymbol{\Sigma}^{-1} \, \mathbf{z}\right|^2}{\mathbf{p}^H \,\boldsymbol{\Sigma}^{-1} \,\mathbf{p}} \gtrless \lambda\, .
\label{MF}
\end{equation}
Under hypotheses $H_0$ and $H_1$, when the Gaussian clutter and thermal noise share the same covariance $\boldsymbol{\Sigma}$, up to a unknown positive scalar factor, the noise is known as partially homogeneous noise. In that case, the optimal detector, invariant to this unknown scale factor in hypothesis $H_0$ leads to the Normalized Matched Filter (NMF) \cite{301849}: 
\begin{equation}
\Lambda_{NMF}(\mathbf{z}) =  \frac{\left|\mathbf{p}^H \,\boldsymbol{\Sigma}^{-1} \,\mathbf{z}\right|^2}{\left(\mathbf{p}^H \,\boldsymbol{\Sigma}^{-1} \,\mathbf{p}\right) \left(\mathbf{z}^H \,\boldsymbol{\Sigma}^{-1} \, \mathbf{z}\right)} \gtrless \lambda\, ,
\label{NMF}
\end{equation}
When the covariance matrix is unknown and the noise is Gaussian, a solution consists in designing two-step adaptive detectors, known as Adaptive Matched Filter (AMF-SCM \cite{Robey1992ACA}) or Adaptive Normalized Matched Filter (ANMF-SCM)  \cite{Kraut2001AdaptiveSD}, by respectively replacing in \eqref{MF} and \eqref{NMF} the true covariance $\widehat{\boldsymbol{\Sigma}}$ by $\widehat{\boldsymbol{\Sigma}}_{SCM}=\displaystyle \frac{1}{K}\sum_{k=1}^K \mathbf{z}_k\, \mathbf{z}_k^H$, the well-known Sample Covariance Matrix (SCM), calculated on a set of independent secondary data: $\mathbf{z}_k=\mathbf{c}_k + \mathbf{n}_k$ for $k \in \{1, \ldots, K\}$. 

In the non-Gaussian case, these two adaptive detectors suffer from regulating the false alarm rate and their detection performance are degraded. 
In a compound Gaussian clutter environment ($\mathbf{c}_k = \sqrt{\tau}_k \, \mathbf{g}_k$), some alternatives consist of designing the so-called the two-step Tyler Adaptive Normalized Matched Filter (ANMF-FP) where the covariance $\boldsymbol{\Sigma}$ is replaced in \eqref{NMF} by the Tyler covariance matrix estimate $\widehat{\boldsymbol{\Sigma}}_{FP} = \displaystyle \frac{m}{K} \sum_{k=1}^K \frac{\mathbf{z}_k^H \, \mathbf{z}_k}{\mathbf{z}_k^H \, \widehat{\boldsymbol{\Sigma}}_{FP}^{-1} \, \mathbf{z}_k}$ built on secondary data $\mathbf{z}_k$'s \cite{tyler1987, pascal08, 6263313}.  The NMF (benchmark) and ANMF-FP are particularly effective \cite{7383755, Pascal8} in scenarii with strong impulsive clutter, providing robustness and texture invariance under $H_0$ hypothesis, where optimal Gaussian detectors (MF, AMF-SCM, ANMF-SCM) are often impractical. Nevertheless, these detectors can suffer from the presence of additive thermal noise which destroys the texture invariance of Tyler's estimator and, in fact, no longer makes it possible to regulate the false alarm (CFAR property). In that case, optimal solutions do not exist and one needs to analyze other approaches. Note that in the case of Gaussian clutter plus thermal noise, there is no issue with MF, AMF-SCM or ANMF-SCM as the sum of independent Gaussian noise vectors is still Gaussian. In the sequel, we propose a VAE detector that can help deal with complex clutter environments with additive thermal noise.

\section{Proposed OOD VAE Detector}
\label{sec:deeplearningdetect}

To overcome the statistical model of clutter plus thermal noise, VAEs can nowadays learn the distribution under $H_0$ by giving a target-free training set $\mathcal{D}_{H_0} = \{\mathbf{z}_1, \cdots,\mathbf{z}_N \in H_0\}$. Then, OOD detection methods  \cite{yang2024} aim to identify whether a sample deviates significantly from the distribution of $\mathcal{D}_{H_0}$, which offers a flexible and generalizable solution for complex and heterogeneous radar environments. By learning the characteristics of In-Distribution (ID) data, OOD detectors can flag samples outside this distribution as potential anomalies. This approach does not rely on target data,  making it well-suited for radar applications where targets are few and unlabeled.

\begin{figure}[htb]
\centering
\includegraphics[width=1.\columnwidth]{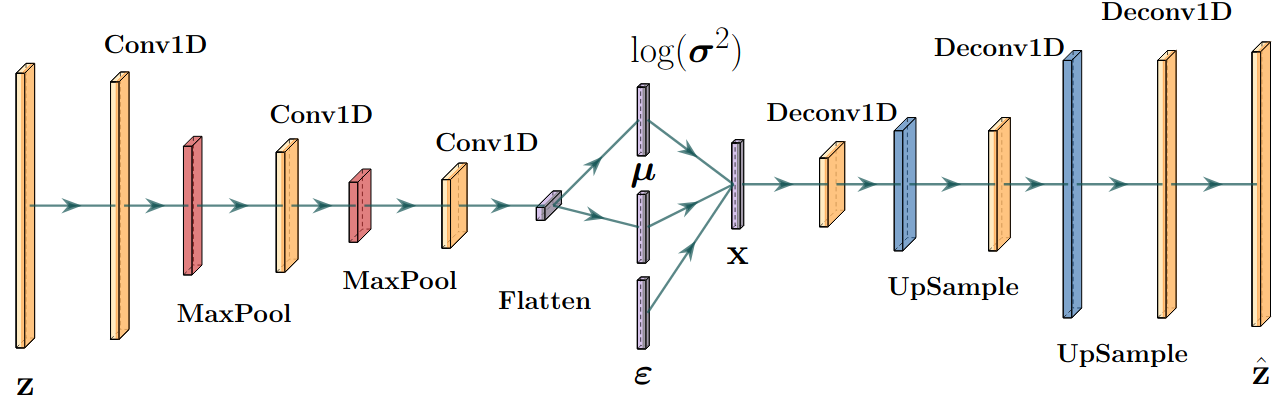}
\caption{VAE network architecture}
\label{fig:vae}
\end{figure}

\begin{figure*}[htb]
\centering
\begin{subfigure}{0.66\columnwidth}
\includegraphics[width=0.99\linewidth, trim={15mm 0mm 28mm 17mm},clip]{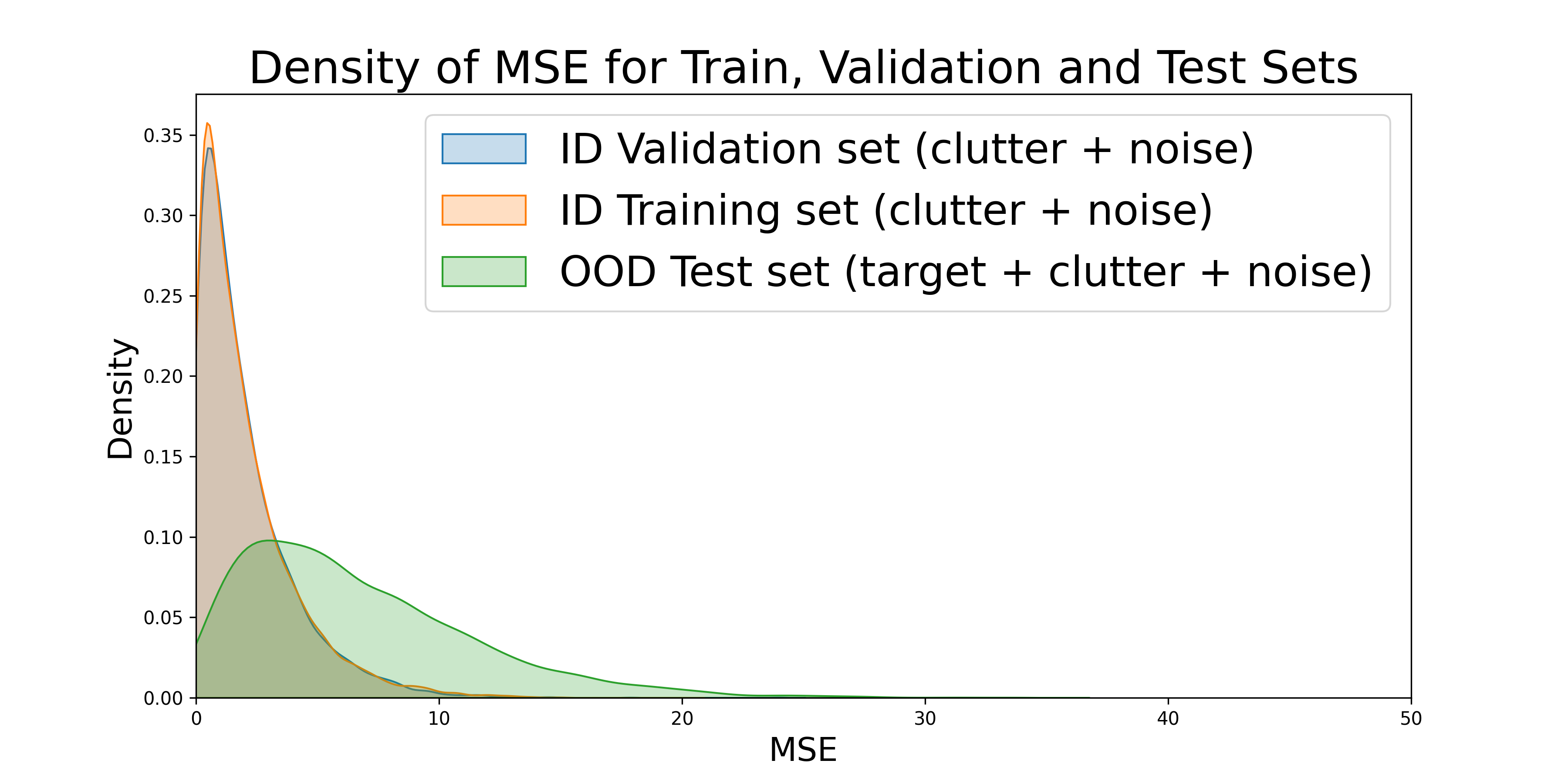}
\caption{SNR = $5$dB}
\end{subfigure}
\hfill\begin{subfigure}{0.66\columnwidth}
\includegraphics[width=0.99\linewidth, trim={15mm 0mm 28mm 17mm},clip]{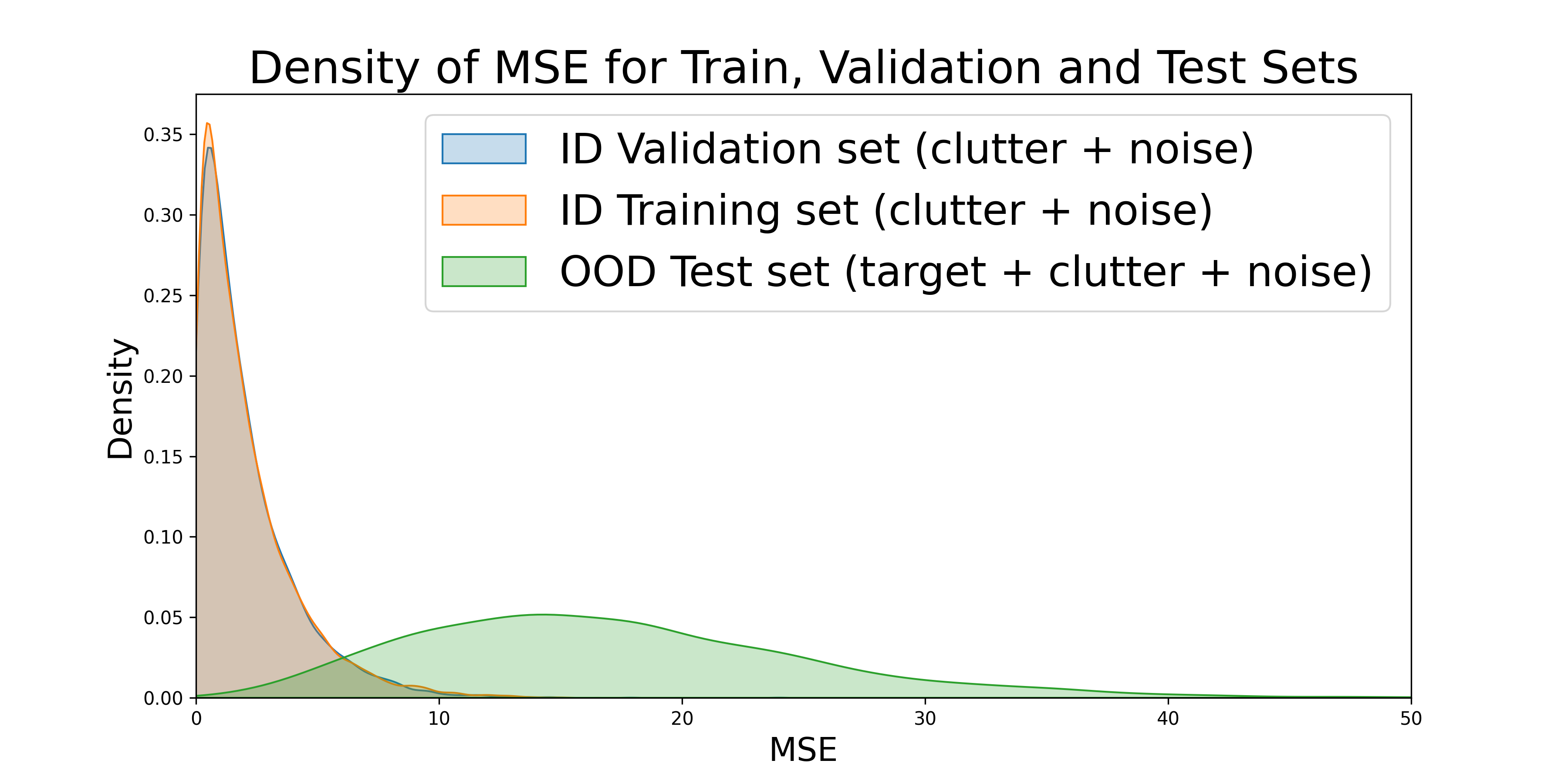}
\caption{SNR = $10$dB}
\end{subfigure}
\hfill\begin{subfigure}{0.66\columnwidth}
\includegraphics[width=0.99\linewidth, trim={15mm 0mm 28mm 17mm},clip]{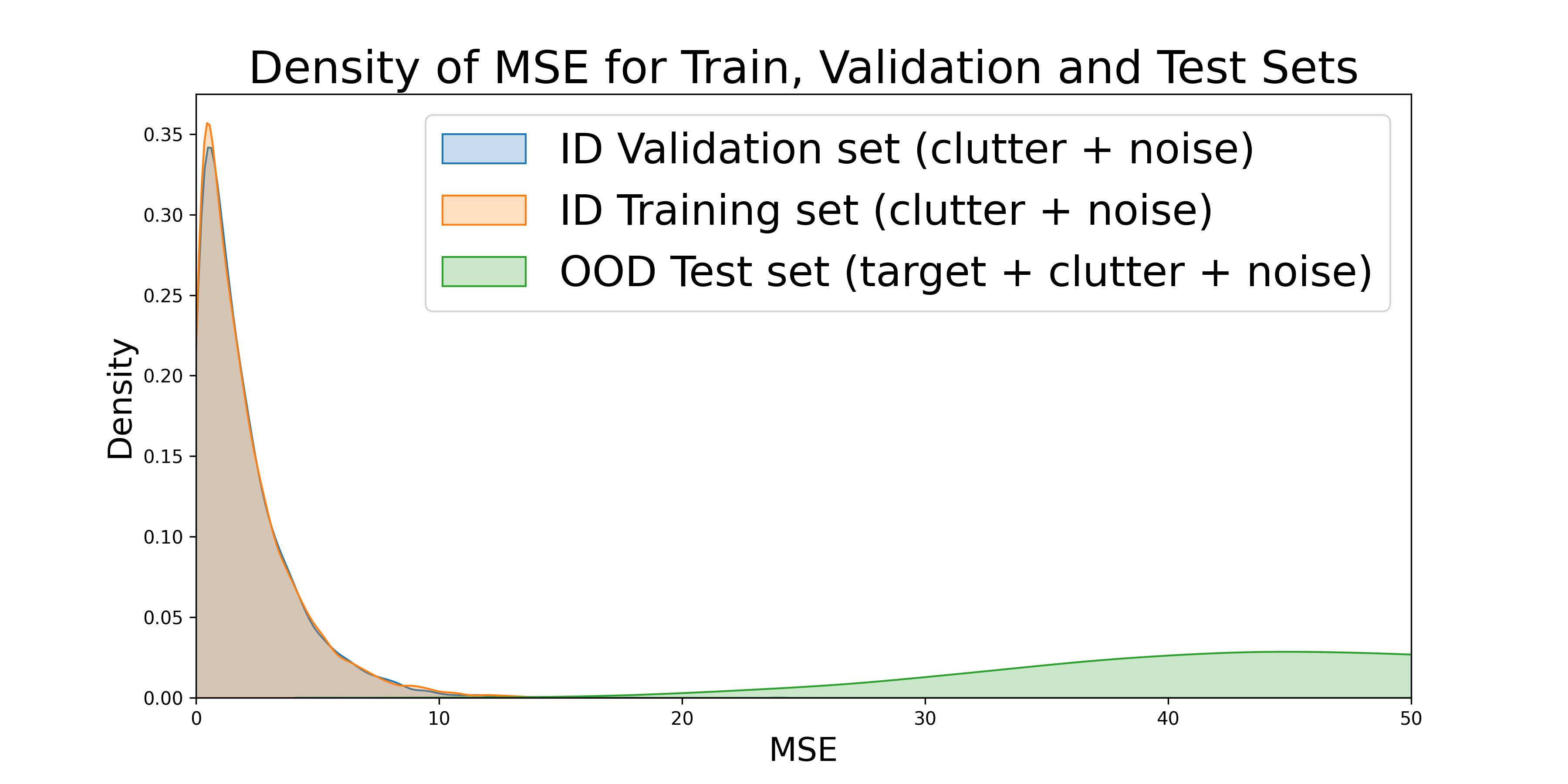}
\caption{SNR = $15$dB}
\end{subfigure}

\caption{$\mathcal{L}_{\text{rec}}$ histogram across SNRs for cCGN + AWGN scenario.}
\label{figmse}
\end{figure*}

\subsection{Proposed VAE Architecture}
\label{ssec:ourvae}

Our VAE architecture is designed to handle the complexity of radar signals while efficiently processing and reconstructing 1D radar signal profiles. The architecture consists of an encoder and decoder, each incorporating convolutional layers to extract and represent the key features of the radar data in a compact latent space.

The encoder (Fig.~\ref{fig:vae}) compresses the high-dimensional radar signals into a lower-dimensional latent space through a series of convolutional blocks. These blocks include convolutional layers, batch normalization, non-linear activation functions, and max-pooling layers to progressively extract and downsample the relevant features. The final output is then passed through fully connected layers to generate the mean $\boldsymbol{\mu} \in \mathbb{R}^q$ and log-variance vectors $\log(\boldsymbol{\sigma}^2) \in \mathbb{R}^q$ which are the parameters of the Gaussian prior distribution. It is well known that a critical aspect of the VAE is the \textit{reparameterization trick} \cite{Kingma_2014}, which allows for the backpropagation of gradients through the stochastic latent variables. Specifically, the generated latent sample $\mathbf{x} = \boldsymbol{\mu} + \boldsymbol{\sigma} \odot \boldsymbol{\epsilon}$ where $\boldsymbol{\epsilon} \sim \mathcal{N}(\boldsymbol{0}, \mathbf{I})$. This process normalizes the latent space variability and ensures that the sampling process is differentiable, enabling the VAE to learn meaningful and continuous latent representations during training.

The decoder (Figure~\ref{fig:vae}) reverses this process, using transposed convolutions and upsampling layers to reconstruct the radar signal from the latent representation $\mathbf{x}$. The output layer applies a convolutional operation to generate the final reconstructed radar signal  $\hat{\mathbf{z}} $.

Our VAE is trained by maximizing the so-called evidence lower bound \cite[Section 2.2]{Kingma_2019}. When the prior and the approximate posterior are Gaussian, this training loss becomes to minimize $\mathcal{L}_{\text{VAE}} = \mathcal{L}_{\text{rec}} + \beta \, \mathcal{L}_{\text{KL}}$ that combines the Mean Square Error (MSE)  $\mathcal{L}_{\text{rec}}(\mathbf{z}, \mathbf{\hat{z}}) = \|\mathbf{z} - \mathbf{\hat{z}}\|^2\ $ for data reconstruction and $\mathcal{L}_{\text{KL}} = -\displaystyle \frac{1}{2} \sum_{i=1}^{q} \left( 1 + \log(\sigma_i^2) - \mu_i^2 - \sigma_i^2 \right)$, the Kullback-Leibler divergence, for the regularization of the latent space where $\mu_i$ and $\sigma_i$ are respectively $i^{th}$ element of $\boldsymbol{\mu}$ and $\boldsymbol{\sigma}$. The hyperparameter \(\beta\) controls the trade-off between these two components. %\cite{burgess2018understandingdisentanglingbetavae}. %This loss function is crucial as it not only guides the VAE in learning how to reconstruct the input data accurately but also in organizing the latent space to ensure that the model generalizes well to unseen data, which is essential for reliable anomaly detection.

\subsection{Detection Strategy and Regulation of PFA}
\label{ssec:detectstrat}

During inference, the VAE processes radar data containing potential targets. Since the VAE was trained exclusively on noise-only data, it struggles to reconstruct signals with targets, resulting in higher MSE when a target is present (see Fig.~\ref{figmse}). Thus, the detection test is $\mathcal{L}_{\text{rec}} (\mathbf{z}, \mathbf{\hat{z}}) \gtrless \lambda_{\text{VAE}}$  where the threshold is calibrated using an evaluation dataset, which also consists solely of clutter plus noise data distinct from the training set. The threshold $\lambda_{\text{VAE}}$ is chosen to regulate the PFA. By analyzing the MSE distribution on the evaluation noise-only dataset, we determine a threshold that maintains the PFA \cite{DISKIN2024109543} at an acceptable level. %Such an approach allows for reliable detection even in challenging radar environments where noise characteristics may vary.
This detection strategy takes advantage of the VAE's ability to model complex distributions, thereby achieving robust target detection while maintaining strict control over the PFA, which is crucial for maintaining system reliability in highly variable radar environments.

\section{Results and Simulations}
\label{sec:results}

In this section, we evaluate the detection performance of the 1D VAE against classical radar detectors: MF, NMF, ANMF-SCM, ANMF-FP, and AMF-SCM. Performance is assessed across various noise scenarii:  correlated Compound Gaussian Noise (cCGN), correlated Gaussian Noise with Additive White Gaussian Noise (cGN + AWGN), and correlated Compound Gaussian Noise with Additive White Gaussian Noise (cCGN + AWGN). The detection performance is measured using Probability of Detection $P_d$ as a function of SNR, with fixed $P_{fa} = 10^{-2}$.

\subsection{Signal and noise characteristics}

To simulate the target, we modeled the echo amplitude $\alpha = \sqrt{\text{SNR}} \, e^{2j\pi \phi} / \sqrt{m}$ where $\phi\in[0,1]$ and the steering vector $\mathbf{p} = \left[1, e^{2j\pi d/m}, \ldots, e^{2j\pi d(m-1)/m} \right]^T$ for $m = 16$ bins, where $d\in \{0, \ldots, m-1\} $ is the $(d+1)$th Doppler bin. The noise and disturbance parameters in the simulation are modeled with $\boldsymbol{\Sigma}_c = \boldsymbol{\mathcal{T}}(\rho)$, with $\rho = 0.5$, textures $\tau$ and $\tau_k$ are sampled according a Gamma distribution $\Gamma(\mu, 1/\mu)$ with $\mu=1$. For adaptive detectors, the covariance matrix is estimated using SCM and  Tyler with $K=2\, m$ secondary data.

\begin{figure*}[htb]
\centering
\begin{subfigure}{0.66\columnwidth}
\includegraphics[width=\linewidth, trim={5mm 3mm 16mm 12mm},clip]{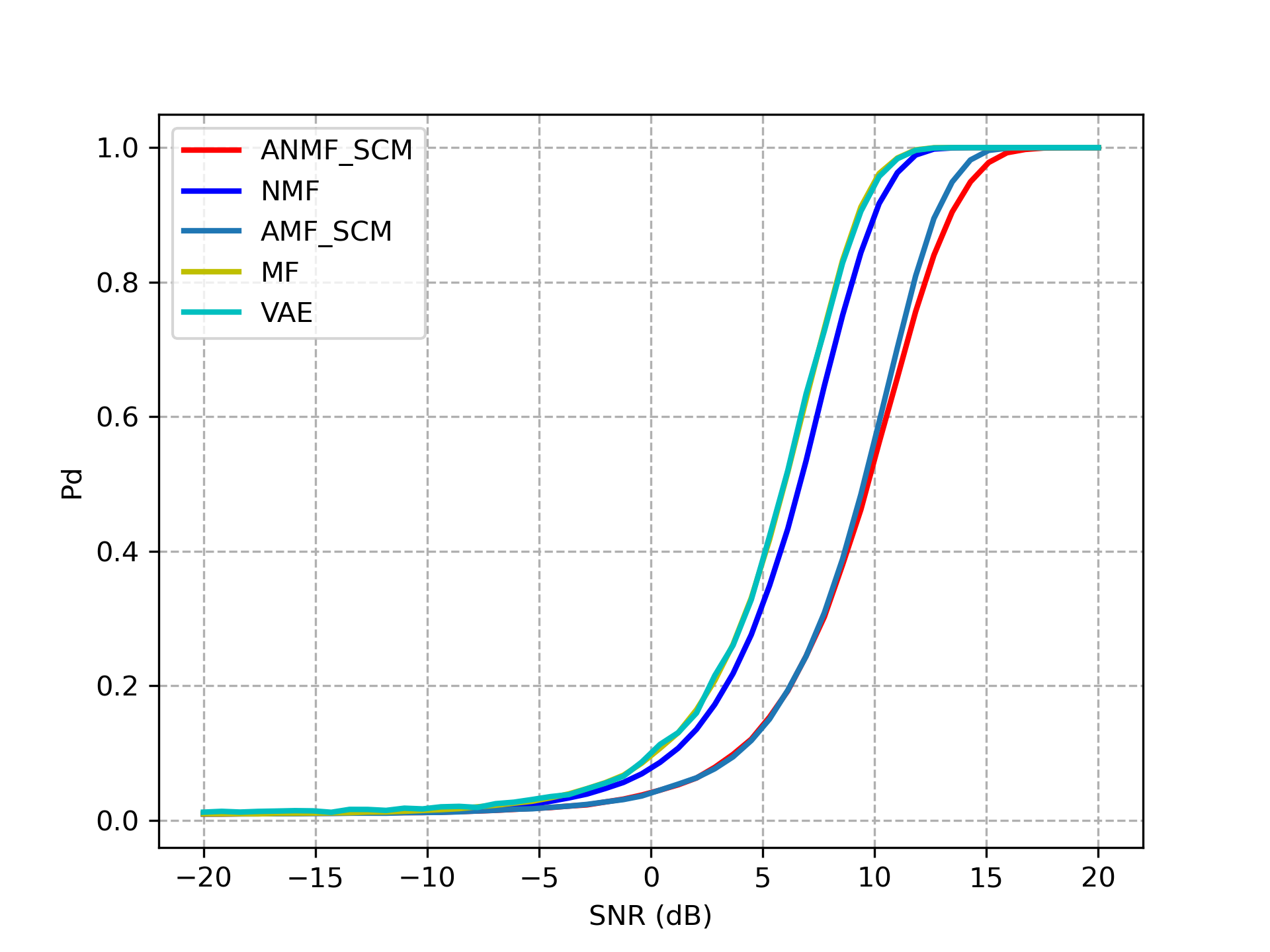}
\caption{cGN + AWGN}
\end{subfigure}
\hfill \begin{subfigure}{0.66\columnwidth}
\includegraphics[width=\linewidth, trim={5mm 3mm 16mm 12mm},clip]{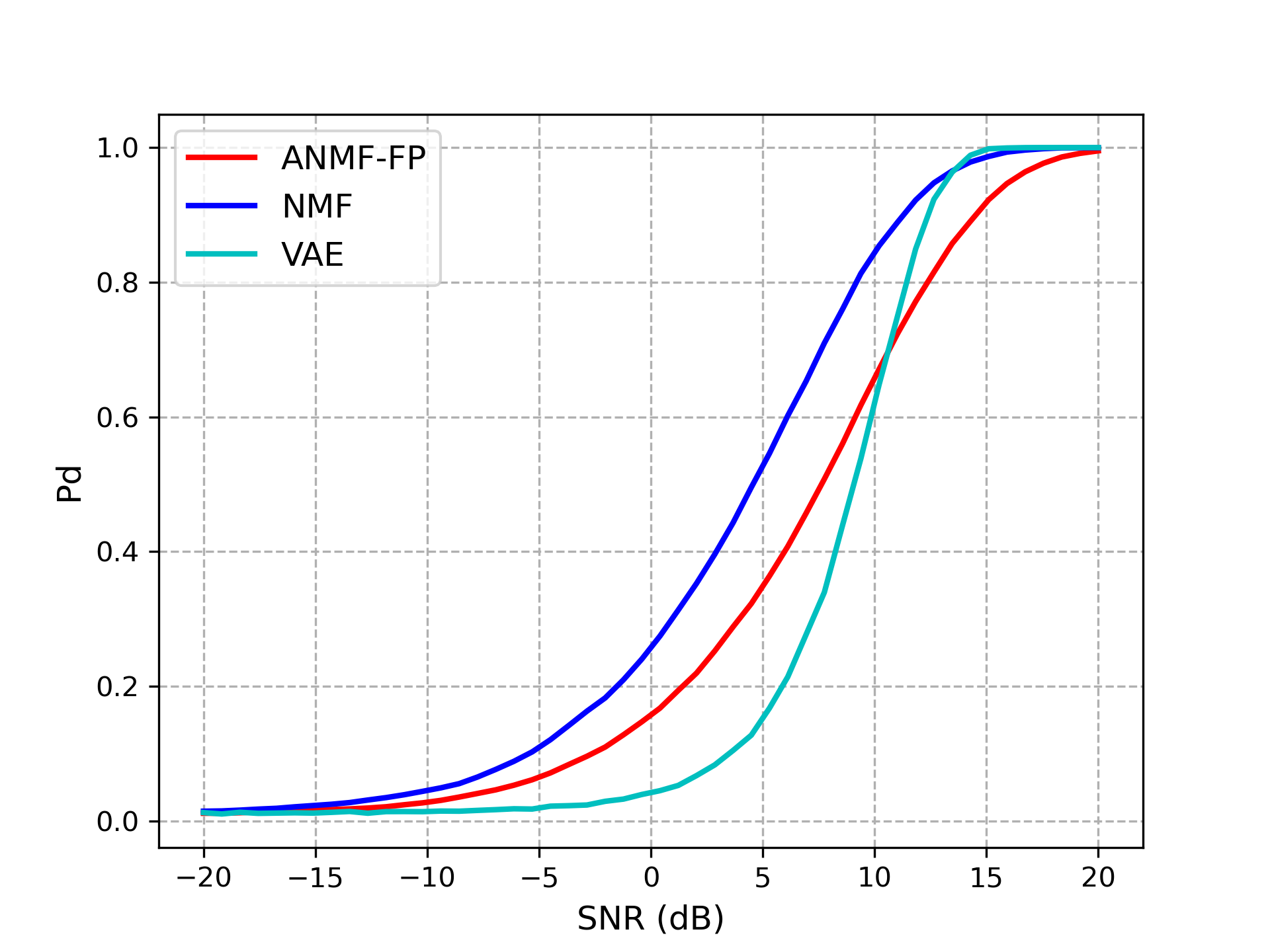}
\caption{cCGN}
\end{subfigure}
\hfill \begin{subfigure}{0.66\columnwidth}
\includegraphics[width=\linewidth, trim={5mm 3mm 16mm 12mm},clip]{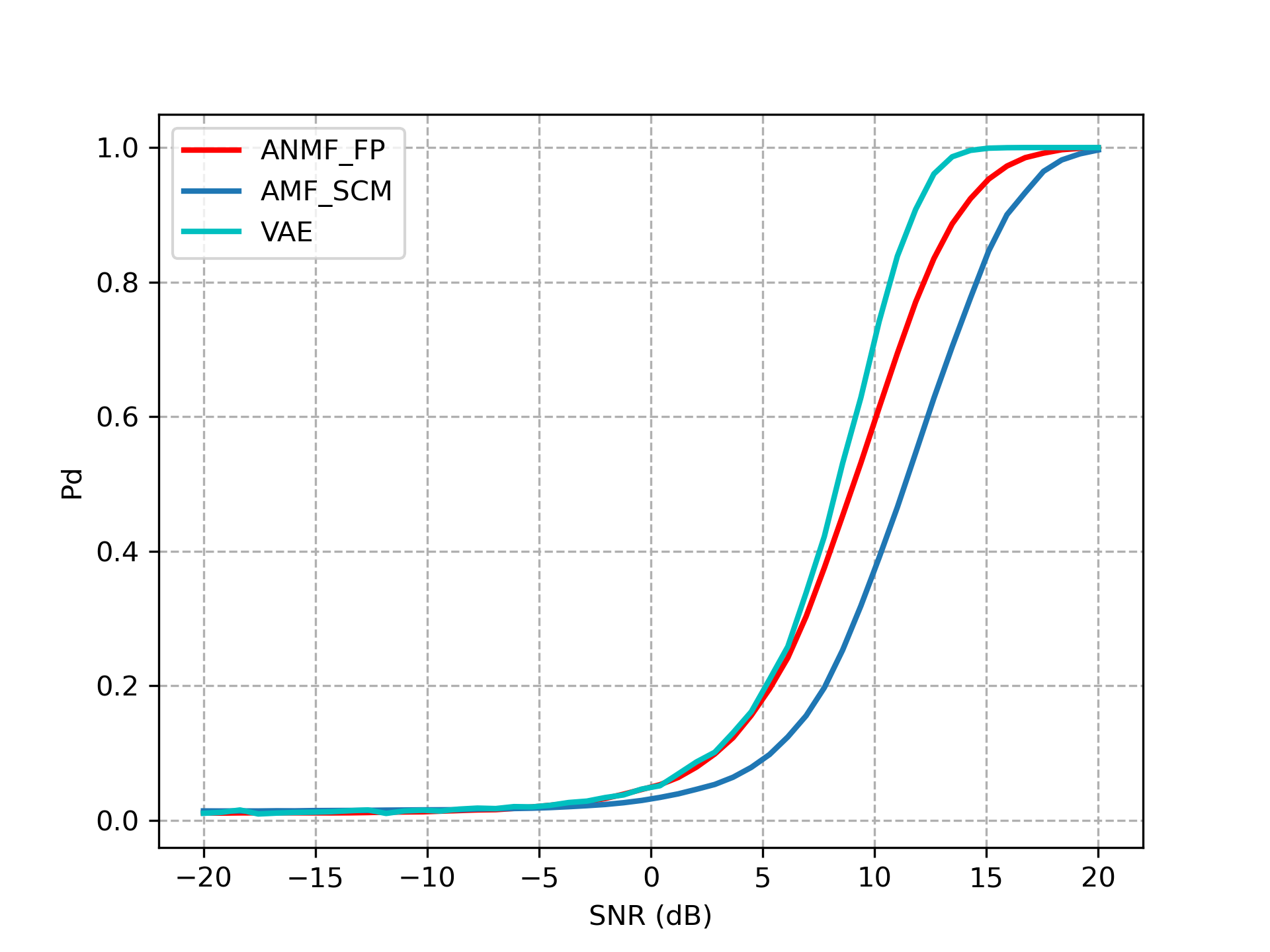}
\caption{cCGN + AWGN}
\end{subfigure}

\caption{Detection performance under different noise for Doppler bin $d=0$ ($P_{fa}= 10^{-2}$, $\rho=0.5$, $\mu=1$, $m=16$, $K=32$).}
 \label{fig:cpdsnr}
\end{figure*}

%The signal and noise were both treated as complex-valued data, consistent with real-world radar signal processing scenarios. 
The analysis is further subdivided into two sections: one focuses on the target embedded in clutter Doppler bin $d=0$ corresponding to the most challenging case. The second is covering all Doppler bins for a comprehensive understanding of Doppler shift influence on the detection performance.

\begin{figure}[htb]
\centering
\begin{subfigure}{0.98\columnwidth}
\includegraphics[width=0.98\linewidth, trim={40mm 3mm 35mm 8mm}, clip]{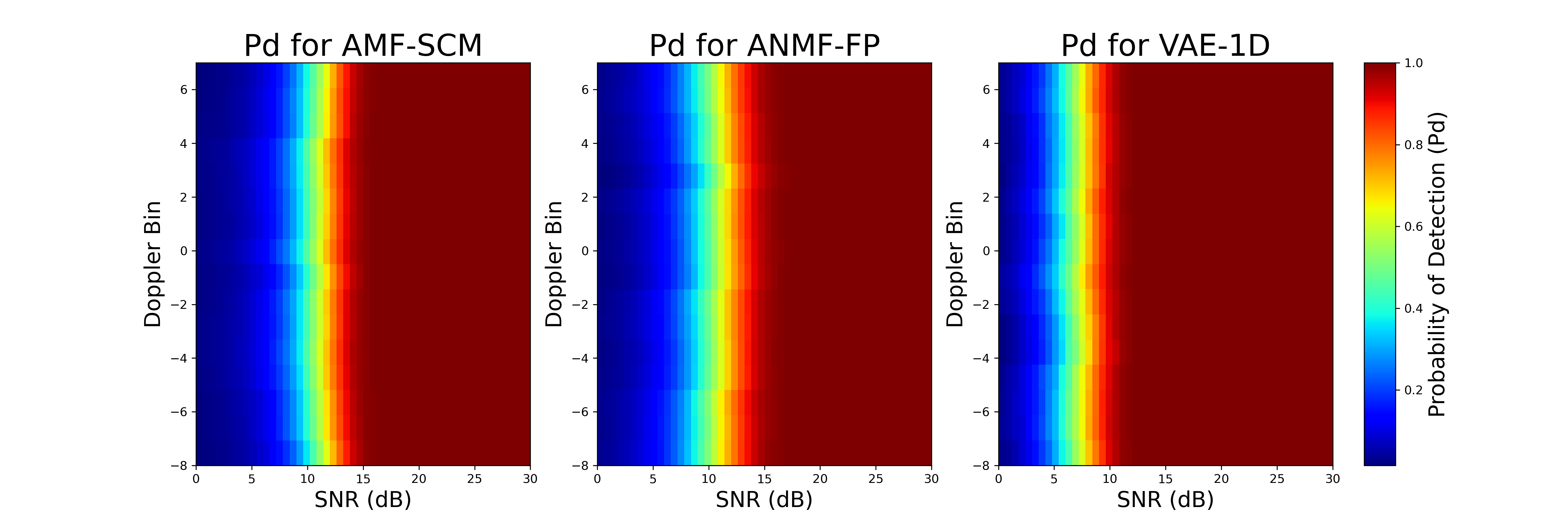}
\caption{cGN + AWGN}
\end{subfigure}
\hfill \begin{subfigure}{0.98\columnwidth}
\includegraphics[width=0.98\linewidth, trim={40mm 3mm 35mm 8mm}, clip]{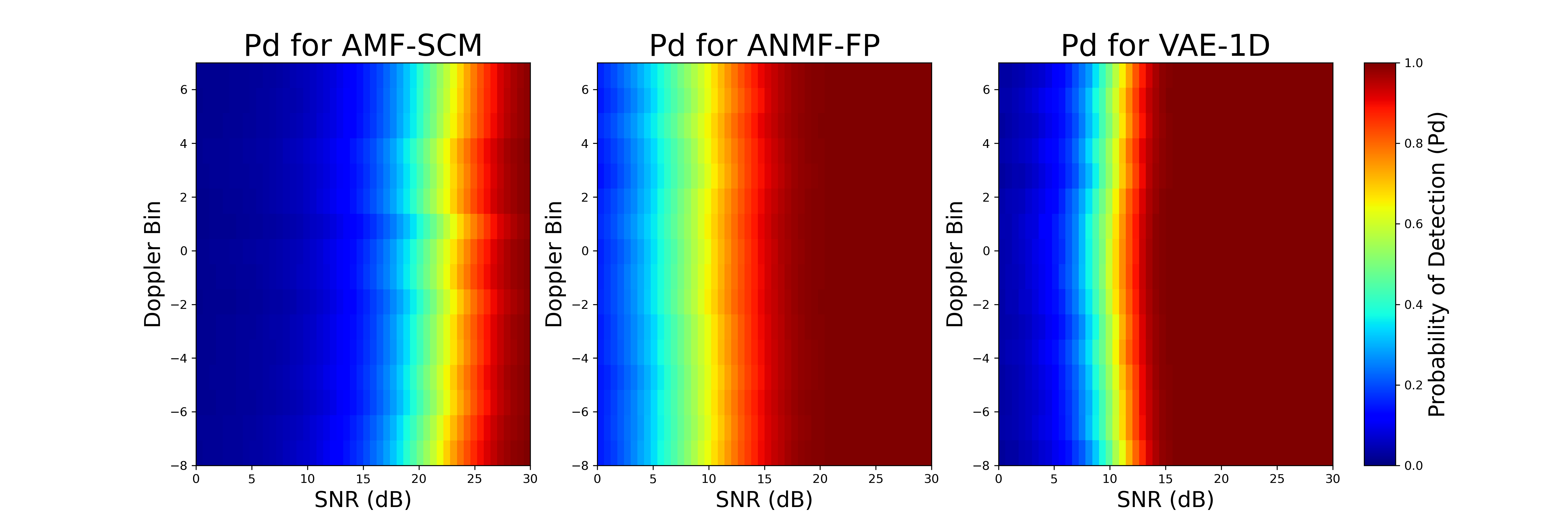}
\caption{cCGN}
\end{subfigure}
\hfill \begin{subfigure}{0.98\columnwidth}
\includegraphics[width=0.98\linewidth, trim={40mm 3mm 35mm 8mm}, clip]{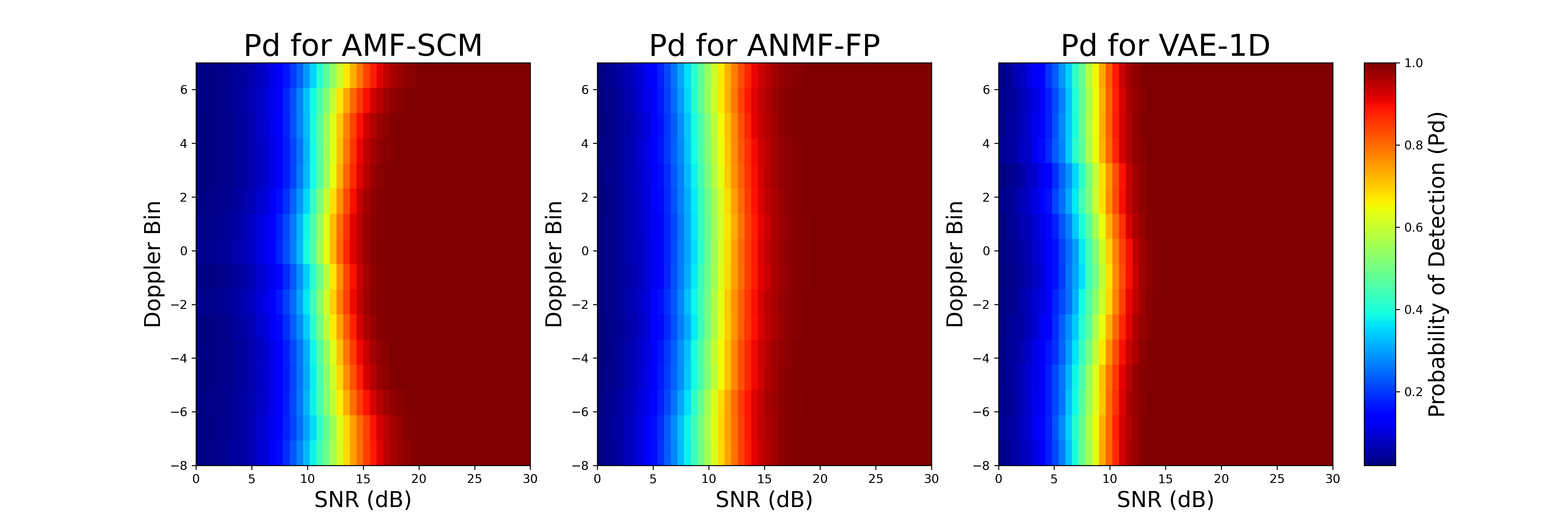}
\caption{cCGN + AWGN}
\end{subfigure}
\caption{$P_d$-SNR-Doppler bin map comparing VAE to AMF and ANMF, under different noise scenarii.}
\label{fig:mpdsnr}
\end{figure}

\subsection{VAE training Setup}
\label{ssec:trainprocess}

For each scenario mentioned above, the VAE is trained on clutter plus noise Doppler profiles. The dataset $\mathcal{D}_{H_0}$ contains $N=15 000$ samples split into $2/3$ for the training and $1/3$ for the validation. Training is conducted over $50$ epochs using the Adam optimizer \cite{Adam}, with a learning rate of \(10^{-3}\). The training loss function is $\mathcal{L}_{\text{VAE}}$ computed with $\alpha = 10^2$. The dimension of the latent space is $12$. One trained, the detection test is based on $\mathcal{L}_{\text{rec}}$ where the $P_{fa}$ is set to $10^{-2}$ computed from an evaluation dataset containing $5000$ samples independently generated from the training set as explained in Section \ref{ssec:detectstrat}. As illustrated in Fig.~\ref{figmse},   $\mathcal{L}_{\text{rec}}$ is a good candidate to separate ID and OOD samples when the SNR increases. 

\subsection{SNR vs $P_d$ analysis for zero Doppler bin}

Fig.~\ref{fig:cpdsnr}-(a) presents the detection performance for zero Doppler bin $d=0$ for cGN + AWGN case. 
The VAE shows strong performance, superposing with the MF detector and outperforming the adaptive detectors like ANMF-SCM and AMF-SCM. This suggests that while the VAE may not exceed classical detectors at higher SNRs, it is a capable contender when handling Gaussian noise environments. In purely cCGN (no added thermal noise), Fig.~\ref{fig:cpdsnr}-(b) reveals that VAE is competitive but behind NMF and ANMF-FP at lower SNRs. However, as the SNR increases, the gap narrows, with VAE aligning more closely with the NMF. This demonstrates that while VAE may struggle initially, its performance improves significantly in less challenging noise conditions. In cCGN + AWGN scenario (Fig.~\ref{fig:cpdsnr}-(c)), the VAE stands out by outperforming adaptive detectors such as ANMF-FP and AMF-SCM. This highlights the VAE's adaptability, as it excels in highly complex noise environments, making it a promising alternative to classical detectors.

\subsection{SNR vs $P_d$ Analysis for all Doppler bins}

Moving to the analysis across all Doppler bins, the correlated Gaussian noise and thermal noise results (Fig.~\ref{fig:mpdsnr}-(a)) show that VAE performs well whatever the Doppler bins and outperforms adaptive detectors like ANMF-FP and AMF-SCM demonstrating its insensitivity to noise characteristics and potential advantages in certain situations.

\begin{comment}
. Although MF and NMF still lead overall, the VAE outperforms adaptive detectors like ANMF-FP and AMF-SCM in specific Doppler shifts, demonstrating its sensitivity to noise characteristics and potential advantages in certain situations.
\end{comment}

For compound Gaussian noise, as shown in Fig.~\ref{fig:mpdsnr}-(b), the VAE competes closely with ANMF-FP at mid to high SNR values and outperforms the AMF-SCM across all Doppler bins, particularly in challenging conditions. This analysis suggests that the VAE offers a robust alternative, especially when operating in less favorable environments.

Finally, under compound Gaussian noise and thermal noise environment, Fig.~\ref{fig:mpdsnr}-(c) illustrates that VAE significantly outperforms adaptive detectors across all Doppler bins. This robust performance, particularly in highly complex noise scenarii, highlights the VAE’s effectiveness in environments where traditional detectors may falter.

\section{Conclusion}
\label{sec:conclu}

The results presented in this study demonstrate the effectiveness of the VAE in radar target detection under various noise conditions. The VAE consistently outperforms traditional detection methods, particularly in scenarii involving impulsive and correlated noise. Its superior performance can be attributed to its ability to model intricate data distributions, making it more robust in detecting radar targets in challenging environments.

\bibliographystyle{IEEEbib}
\bibliography{refs}

\end{document}